# NONPARAMETRIC CURVE ALIGNMENT

*Marwan A. Mattar*\*, *Michael G. Ross*\*\*, *Erik G. Learned-Miller*\*

\* Department of Computer Science, University of Massachusetts, Amherst, MA 01002
\*\* Department of Brain and Cognitive Sciences, MIT, Cambridge, MA 02139

## ABSTRACT

Congealing is a flexible nonparametric data-driven framework for the joint alignment of data. It has been successfully applied to the joint alignment of binary images of digits, binary images of object silhouettes, grayscale MRI images, color images of cars and faces, and 3D brain volumes. This research enhances congealing to practically and effectively apply it to curve data. We develop a parameterized set of nonlinear transformations that allow us to apply congealing to this type of data. We present positive results on aligning synthetic and real curve data sets and conclude with a discussion on extending this work to simultaneous alignment and clustering.

## 1. INTRODUCTION

Joint alignment is the operation which transforms data to increase an application-specific measure of their mutual similarity. Its purpose is to remove unwanted variability in the data. For example, multiple realizations of any stochastic process contain variations in time and amplitude. The resulting curves [1] can be aligned to remove this variability by allowing certain transformations on the data. This is useful for recovering the underlying data needed for a task (e.g. for speech processing [2, 3]) or for obtaining a representation invariant to unmodeled factors (e.g. for dimensionality reduction [4]).

While several models have been proposed for alignment of curve data [3, 5, 6, 2], all make significant assumptions about their form (a single underlying curve [3, 5] or parametric assumptions [2]) or the transformations they can undergo [2, 6]. On the other hand, congealing is a nonparametric framework for joint alignment that makes very few assumptions about the data and can accommodate any continuous family of transformations, making it more widely applicable.

Congealing iteratively optimizes a set of transformation parameters (associated with the data) using an information-theoretic objective function as a measure of joint alignment. Each data point (in our case a single curve, $C_k$) is associated with a transformation parameter, $V_k$. Congealing *simultaneously* searches over the space of all the data points' transformation parameters to find those that maximize a measure of the *joint* alignment. More formally, when aligning $N$ curves, congealing searches for the transformation parameters $\{V_1, ..., V_N\}$ that maximizes a measure of the transformed curves, $\mathcal{S}(\{\mathcal{T}(C_1, V_1), ..., \mathcal{T}(C_N, V_N)\})$, where $\mathcal{T}(\cdot, \cdot)$ is the transformation function and $\mathcal{S}(\{\cdot\})$ measures the joint alignment of a set of curves.

The freedom to use any set of transformations, transformation parameterization, optimization procedure and similarity function makes congealing more of a framework than a specific algorithm. Different choices of these factors result in different algorithms (e.g. [7, 8, 9, 10]). This paper derives a congealing algorithm for the joint alignment of curve data by specifying a parameterized set of useful nonlinear alignment transformations. Section 2 describes the algorithm, Section 3 presents positive results on aligning synthetic and real curve data sets and Section 4 analyzes the algorithm's performance and concludes with a discussion on extending it to simultaneous alignment and clustering.

## 2. ALGORITHM

In this section we address each of the congealing design choices and present what we found to work well in practice. Although we present congealing as an optimization framework, it has an intuitive probabilistic interpretation [7, 1].

**Allowable transformations.** The algorithm allows nonlinear scaling in time, linear scaling in amplitude and amplitude translation. Nonlinear scaling implies that different regions in a curve are scaled by different amounts. Because we assume the curves have the same length,[2] linear time scaling and time translation are invalid transformations. In Section 4 we consider nonlinear amplitude scaling.

**Measure of joint alignment.** The algorithm measures joint alignment using the sum of location-wise differential entropies. Each time step's entropy is independently calculated,

---

Supported by the National Science Foundation under grants ATM 0325167 and CAREER CSE IIS 0546666. See [1] for a more detailed report.

[1]Here, a curve is simply any 1-D function and may be alternatively referred to as *time series* or *signal*.

[2]This is a common assumption for curve alignment and can easily be relaxed in our algorithm to allow translation and/or linear scaling in time.

treating the set of values from all curves at each time step as samples from a probability density function. Those samples are used to compute the entropy at each time step and then are summed across all the time steps to estimate the total entropy of the data. Ideally the algorithm would compute the joint entropy of all curves by treating each one as a sample from an $M$-dimensional density, where $M$ is the curve length. However, this estimation problem is infeasible because most curve data sets have more time steps than curves. Instead, the algorithm adds entropies at each time step, which corresponds to an implicit assumption that the time steps are independent after alignment [7]. The entropy of each time step is calculated using the efficient distribution-free Vasicek estimator [11].

Entropy is a measure of spread that does not make any assumptions about the form of the underlying distribution (as opposed to variance, for example, which is more appropriate for a unimodal distribution). By maximizing negative entropy, the algorithm can align multi-modal data sets by automatically discovering the modes via the alignment procedure.

**Optimization procedure.** We found that the standard congealing gradient descent [7] converged quickly and avoided local minima. In each iteration this procedure iterates over all the transformation parameters for all the curves and updates their parameters by a small random amount so as to decrease the sum of entropies. It stops when several consecutive iterations do not change the objective function.

**Transformation parameterization.** One of the biggest difficulties in developing alignment algorithms is in parameterizing the transformations. Some parameterizations are obvious, such as those for linear scaling and translations, but others, such as those for nonlinear time scaling, are more challenging. Therefore, several algorithms, including *dynamic time warping*, do not parameterize nonlinear scaling in time, and attempt to search all possible monotonic scaling functions. Other approaches exclude nonlinear scalings (e.g. [2]) or take a local approach to alignment (e.g. [3]) which can arbitrarily warp the curves. These approaches miss the benefits of good parameterizations, including low dimensionality, high modeling capacity and efficient computation.

Our algorithm allows changes in amplitude of the form $\alpha y + \beta$, where $y$ is the original amplitude. Hence, amplitude can be shifted or scaled linearly. It parameterizes nonlinear time scaling as a monotonic warping function, $h(t)$, that maps the time steps of the original curve to the time steps of the aligned curve. The aligned curve is then generated via bilinear interpolation, $C_{aligned}(t) \approx C_{original}(h(t))$.

There are several ways to parameterize $h(t)$ (e.g. [12]). We found that Ramsay's method [13] using a Fourier basis (as opposed to the recommended b-splines) was the most efficient and compact — only 4 basis functions were needed. Ramsay's method is based on the fact that monotone functions are a family of functions defined by $\frac{\partial^2 h}{\partial t} = w(t)\frac{\partial h}{\partial t}$, where $w(t)$ is an unconstrained coefficient function. This equation has the following solution, $h(t) = \frac{1}{Z}\int_0^t \exp\left(\int_0^{r=t} w(s)\,ds\right) dr$, where $Z$ is the normalizing constant, $Z = \int_0^1 \exp\left(\int_0^1 w(s)\,ds\right)$. Using this equation, the algorithm can calculate a monotone warping function $h(t)$ from any $w(t)$, which represents the relative curvature of $h(t)$. We used a linear combination of sine and cosine functions at varying frequencies to parameterize the coefficient function. Thus $w(t) = \sum_{k=1}^{K} \phi_k \sin(2\pi kt) + \omega_k \cos(2\pi kt)$, where the weights $(\phi_1, ..., \phi_k, \omega_1, ..., \omega_k)$ are the parameters for nonlinear time scaling. Using only two frequencies $(\frac{1}{2}, 1)$ provided sufficient modeling capacity.

**Summary.** These design choices resulted in an efficient and flexible algorithm for the joint alignment of curve data. It employs a search procedure that makes no assumptions about the form of the curves and only weak assumptions about the structure of transformations. Furthermore, each curve's transformation is parameterized by only six parameters.

## 3. EXPERIMENTAL RESULTS

It is difficult to quantitatively evaluate the performance of alignment algorithms and qualitative evaluations are only meaningful if they are performed by an expert who understands the nature of the data set. In practice, the quality of alignment depends on the application. Measuring the sum of squared differences, for example, could be misleading because a collection of curves can always be trivially aligned — by setting them all to zero, for example — and even many non-trivial "perfect" alignments might be removing useful data. Therefore, we performed a series of alignment and classification experiments on synthetic and real data sets to evaluate our algorithm. Using classification results to measure alignment quality allows us to determine if the algorithm is providing a tangible benefit.

**Aligning synthetic data sets.** The purpose of the first series of experiments was to evaluate the effectiveness of the algorithm's parameterization and search techniques using synthetic data sets. We selected five curves from the UCR time-series repository and for each curve created five synthetic data sets of increasing difficulty. Each data set was created by randomly transforming the original curve 50 times. We created three groups of these 25 data sets and in each group only one of the three transformations (linear amplitude scaling, amplitude translation, nonlinear time scaling) was performed. This allowed us to study each transformation independently, and provided us with the ground-truth underlying curves.

We aligned each of the 75 data sets using our algorithm, but in each case we restricted the algorithm to using the type of transformation that created that data set. For these synthetic data experiments we used variance instead of entropy in the objective function. This allowed us to more accurately assess our transformation parameterization given that the data sets are unimodal. In 72 of the 75 experiments, our algorithm perfectly aligned the data set to the original curve that gen-erated it. Figure 1 presents an alignment result from each of three transformations on the most difficult data sets we generated. These results suggest our algorithm has the capacity to align curve data corrupted by large transformations while avoiding local minima.

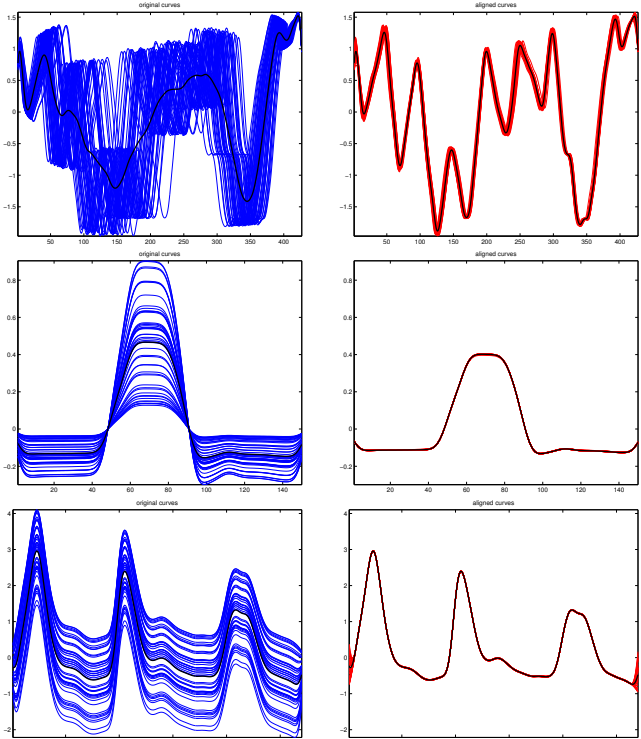

**Fig. 1**. Alignment results on three synthetic data sets. Original curves are in the first column and aligned curves are in the second. The black curve in each plot is the mean curve.

**Curve classification via alignment.** Because the purpose of alignment is to eliminate undesirable variation, the performance of an alignment procedure can be assessed by investigating whether it improves the performance of a classifier [9]. Often, bringing the curves into correspondence simplifies the classification problem and improves the classifier's performance. We performed two types of classification experiments (using entropy in the objective function), the first based on supervised alignment and the second based on unsupervised alignment. We used six data sets from the UCR time-series repository: Beef, Coffee, ECG200, FaceFour, GunPoint, and Trace. In all cases we used a simple K-nearest neighbor classifier ($K = 10$) so that it would be more sensitive to the effects of the data alignment process. The classification accuracies are plotted in Figure 2.

Classification based on supervised alignment involves aligning the (train and test) curves from each category independently and then performing classification. We used the same train and test splits specified in the repository [14]. We compared these results to performing classification without alignment. As Figure 2 shows, the accuracies for aligned data were higher than those for unaligned data. These improvements show that the aligned data still had enough information to distinguish classes and removed irrelevant variation. Figure 3 shows a sample alignment result. **However, it is possible, since alignments were done separately on each class, that the alignment algorithm itself *introduced* information allowing the separation of classes**. To address this issue, we turn to "unsupervised alignment" experiments.

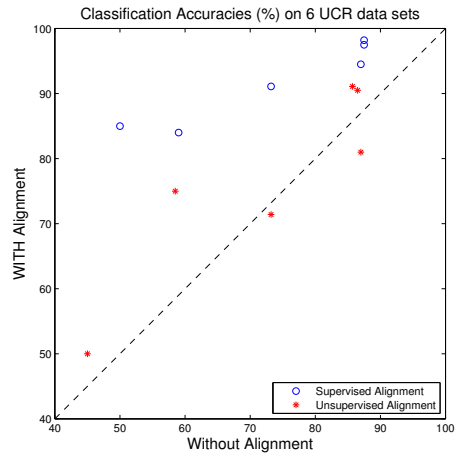

**Fig. 2**. Classification results. Each data point represents an experiment outcome, its abscissa denotes accuracy without alignment and its ordinate denotes accuracy with alignment.

Classification based on unsupervised alignment involves aligning the train and test curves from *all* the categories simultaneously, *without knowing the class label for each curve*, and then performing classification. Hence it represents a more realistic application scenario. In these experiments we only used nonlinear time scaling because our goal was to bring the curves into locational correspondence to simplify classification. We performed 10-fold stratified cross validation and compared the results to performing classification without alignment (using the same folds). As Figure 2 shows, the mean accuracies with alignment were higher than the mean accuracies without alignment in most cases. Figure 4 illustrates the one data set for which alignment substantially harmed classification performance. This data set was a special case in which curve amplitudes were similar across both categories and the most discriminating feature was time displacement. The alignment procedure eliminated this displacement, making the classification problem harder.

## 4. DISCUSSION AND FUTURE WORK

We have presented an efficient joint alignment algorithm for curve data, which demonstrated the utility of an efficient parameterization for nonlinear transformations. We tested our algorithm on a wide range of complex curve data sets and in almost all the cases it improved the correspondence across the curves and improved the performance of a classifier on those data sets. Compared to existing algorithms, it makes fewer assumptions about the distribution of the curves and the transformations, making it more widely applicable.

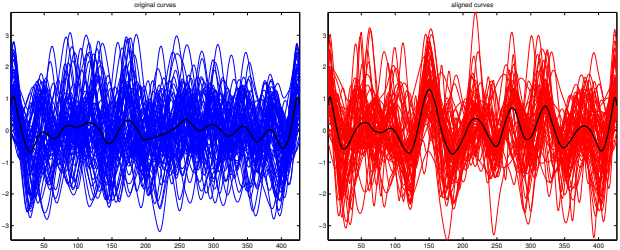

**Fig. 3**. Alignment result on OSULeaf (category 1) data set with mean curves overlaid.

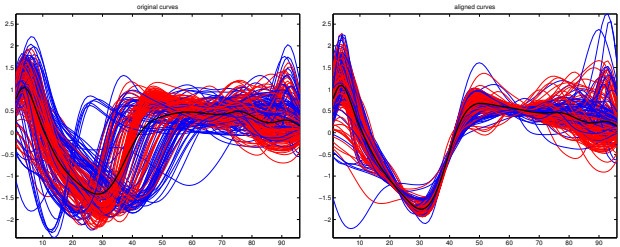

**Fig. 4**. Alignment result on ECG200 (both classes, color-coded) data set.

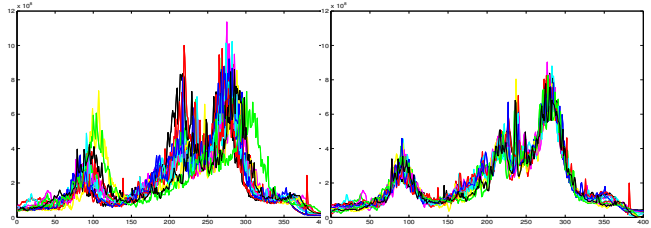

**Fig. 5**. Alignment result on HPLC-MS data [3]. Left: original. Right: aligned.

It is interesting to note that the behavior of our algorithm using a location-wise variance objective function and incorporating nonlinear amplitude scaling is similar to the *continuous profile model* (CPM) [3]. CPM uses an HMM-based alignment procedure that allows each observation to move in time or amplitude and assumes a single underlying curve. Figure 5 shows the alignment result of our slightly modified algorithm on the same data presented in [3]. These alignment results are very similar to those generated by CPM. This highlights a desirable characteristic of our algorithm: It can be easily tuned to handle specific cases of interest. Our software implementation reflects this property.

Future work will include extending the algorithm to simultaneous alignment and clustering. This will allow us to overcome the limitations of the independence assumption in our current algorithm which can fail when presented with complex data sets arising from multiple modes. The use of Dirichlet process priors will allow us to maintain the nonparametric nature of our approach.